\documentclass{article}

\usepackage{etoolbox}
\newtoggle{todo}
\newtoggle{arxiv}
\toggletrue{todo}
\toggletrue{arxiv}

\usepackage{microtype}
\usepackage{graphicx}
\usepackage{caption}
\usepackage{subcaption}
\usepackage{booktabs} 

\usepackage{hyperref}
\usepackage[inline]{enumitem}
\usepackage[table]{xcolor}
\usepackage{hhline}
\usepackage{comment}
\usepackage{bbm}

\iftoggle{arxiv}{
  \usepackage[numbers]{natbib}
  \setlength{\textwidth}{6.5in}
  \setlength{\textheight}{9in}
  \setlength{\oddsidemargin}{0in}
  \setlength{\evensidemargin}{0in}
  \setlength{\topmargin}{-0.5in}
  \newlength{\defbaselineskip}
  \setlength{\defbaselineskip}{\baselineskip}
  \setlength{\marginparwidth}{0.8in}
}{

  \usepackage{natbib}
  \usepackage{nature}
 
}

\usepackage{parskip}
\usepackage{xspace}
\usepackage{amsopn,amsmath,amsthm,amssymb}
\usepackage{algorithm}
\usepackage{algorithmic}
\usepackage{breqn}
\usepackage{tabu}
\usepackage{url}
\usepackage{color}
\usepackage{mathtools}
\usepackage{pbox}
\usepackage{scalerel}

\usepackage{listings}
\definecolor{codegreen}{rgb}{0,0.6,0}
\definecolor{codegray}{rgb}{0.5,0.5,0.5}
\definecolor{codepurple}{rgb}{0.58,0,0.82}
\definecolor{backcolour}{rgb}{1.0,1.0,1.0}

\lstdefinestyle{mystyle}{
    backgroundcolor=\color{backcolour},   
    commentstyle=\color{codegreen},
    keywordstyle=\color{magenta},
    numberstyle=\tiny\color{codegray},
    stringstyle=\color{codepurple},
    basicstyle=\ttfamily\footnotesize,
    breakatwhitespace=false,         
    breaklines=true,                 
    captionpos=b,                    
    keepspaces=true,                                  
    numbersep=5pt,                  
    showspaces=false,                
    showstringspaces=false,
    showtabs=false,                  
    tabsize=2
}

\usepackage{upgreek}

\newcolumntype{N}{>{\centering\arraybackslash}m{.5in}}
\newcolumntype{G}{>{\centering\arraybackslash}m{2in}}
\usepackage[normalem]{ulem}
\useunder{\uline}{\ul}{}

\newcolumntype{C}[1]{>{\centering}m{#1}}

\iftoggle{arxiv}{
  \title{Distilling Large Language Models \\ for Efficient Clinical Information Extraction}
  \usepackage{authblk}
  
  \author[1]{Karthik S. Vedula*\textsuperscript{\textdagger}}
  \author[2]{Annika Gupta*}
  \author[3]{Akshay Swaminathan*}
  \author[3]{\authorcr Ivan Lopez}
  \author[3]{Suhana Bedi}
  \author[4,5,6]{Nigam H. Shah}
  \affil[1]{Poolesville High School, Poolesville, MD, USA}
  \affil[2]{University of California Santa Cruz, Santa Cruz, CA, USA}
  \affil[3]{Department of Biomedical Data Science, Stanford University School of Medicine, Stanford, CA, USA}
  \affil[4]{Department of Medicine, Stanford University School of Medicine, Stanford, CA, USA}
  \affil[5]{Clinical Excellence Research Center, Stanford University School of Medicine, Stanford, CA, USA}
  \affil[6]{Technology and Digital Solutions, Stanford Health Care, Palo Alto, CA, USA}
}{

}

\date{\today}

\begin{document}

\maketitle

\begin{center}
$^{*}$Equal Contribution
\textsuperscript{\textdagger}Corresponding Author: \texttt{karthik@vedula.me}
\end{center}

\section*{Abstract}

\noindent \textbf{Objective}: Large language models (LLMs) excel at clinical information extraction but their computational demands limit practical deployment. Knowledge distillation—the process of transferring knowledge from larger to smaller models—offers a potential solution. We evaluate the performance of distilled BERT models, which are approximately 1,000 times smaller than modern LLMs, for clinical named entity recognition (NER) tasks.

\textbf{Materials and Methods:} We leveraged state-of-the-art LLMs (Gemini and OpenAI models) and medical ontologies (RxNorm and SNOMED) as teacher labelers for medication, disease, and symptom extraction. We applied our approach to over 3,300 clinical notes spanning five publicly available datasets, comparing distilled BERT models against both their teacher labelers and BERT models fine-tuned on human labels. External validation was conducted using clinical notes from the MedAlign dataset.

\textbf{Results:} For disease extraction, F1 scores were 0.82 (teacher model), 0.89 (BioBERT trained on human labels), and 0.84 (BioBERT-distilled). For medication, F1 scores were 0.84 (teacher model), 0.91 (BioBERT-human), and 0.87 (BioBERT-distilled). For symptoms: F1 score of 0.73 (teacher model) and 0.68 (BioBERT-distilled).  Distilled BERT models had faster inference (12x, 4x, 8x faster than GPT-4o, o1-mini, and Gemini Flash respectively) and lower costs (85x, 101x, 2x cheaper than GPT-4o, o1-mini, and Gemini Flash respectively). On the external validation dataset, the distilled BERT model achieved F1 scores of 0.883 (medication), 0.726 (disease), and 0.699 (symptom).

\textbf{Conclusions:} Distilled BERT models were up to 101x cheaper and 12x faster than state-of-the-art LLMs while achieving similar performance on NER tasks. Distillation offers a computationally efficient and scalable alternative to large LLMs for clinical information extraction.

\pagebreak

\section*{Introduction}
Clinical notes in electronic health records contain valuable unstructured information that often isn’t captured in structured fields \cite{ross_big_2014}.  Converting this free-text information into structured data enables cohort selection\cite{wornow_zero-shot_2024}, observational analysis\cite{callahan_research_2020}, and question-answering systems that enhance clinician efficiency.\cite{lamurias_lasigebiotm_2019}  However, extracting information from these clinical notes remains challenging.\cite{zweigenbaum_frontiers_2007}\cite{wang_clinical_2018}  Named entity recognition (NER), which classifies key entities in text into predefined categories like diseases, medications, or symptoms, is an important task in this process.\cite{lample_neural_2016}

Traditional approaches to clinical NER include rule-based methods using string matching and medical ontologies like the Unified Medical Language System (UMLS).\cite{bodenreider_unified_2004}\cite{liao_development_2015}\cite{campillos-llanos_clinical_2021} While these approaches are interpretable and computationally efficient, they often fail to capture the diverse representations of clinical entities, including synonyms, abbreviations, nuanced descriptions, and misspellings.\cite{liao_development_2015}

Machine learning approaches, such as BERT-based models, have demonstrated superior performance.\cite{devlin_bert_2019}\cite{fries_ontology-driven_2021} Domain-specific BERT variants like BioBERT\cite{lee_biobert_2020} and ClinicalBERT\cite{huang_clinicalbert_2020} have been developed to better handle biomedical and clinical terminology.  However, current clinical NER models (fine-tuned BERT Models) tend to be narrowly focused on specific domains or entity types, like radiology, limiting their broad applicability.\cite{chaves_rales_2023} Additionally, fine-tuning requires large amounts of annotated data, which is expensive and time-consuming to produce. Weak supervision using rule-based methods and ontologies—such as TROVE, which generates weak labels from UMLS ontologies to train a BERT-based model for NER—offers one solution.\cite{fries_ontology-driven_2021}

Large language models (LLMs) have demonstrated strong performance in clinical NER tasks through zero-shot or few-shot prompting, reducing the need for extensive labeled data.\cite{monajatipoor_llms_2024} However, these models require significant computational resources for local deployment and can be costly.\cite{noauthor_pricing_nodate} Additionally, proprietary LLMs often require HIPAA-compliant endpoints to handle protected health information (PHI), which further complicates their deployment in healthcare settings. These challenges highlight the need for more efficient and compliant solutions in the healthcare domain.

Knowledge distillation offers a promising solution to these challenges. This technique transfers knowledge from larger models to smaller ones, potentially addressing the limitations of both domain-specific BERT models and computationally expensive LLMs.\cite{hinton_distilling_2015} Recent studies have demonstrated successful distillation from large models such as GPT-4 into medium-sized LLMs such as LLaMA\cite{zhou_universalner_2023}, and from BERT-based models to even smaller architectures.\cite{rhouma_leveraging_2024} In the medical domain, distilled models have achieved impressive results — DistilFLERT and distilled PubMedBERT models have shown success in various medical applications.\cite{rhouma_leveraging_2024}\cite{gu_distilling_2023} 

However, existing approaches have several limitations. First, they typically focus on single note type (e.g., discharge summaries) or single entity type (e.g., medications only), limiting their practical utility across diverse clinical settings. Second, prior work has not rigorously investigated the generalizability of distilled models through external validation using notes from different health systems and note types. Third, existing approaches rely on single teacher models rather than exploring the potential benefits of combining multiple teacher labelers that leverage both LLMs and medical ontologies. This gap is particularly significant given that different teacher labelers may capture complementary aspects of clinical entities, potentially improving the robustness and accuracy of the distilled models.
In this paper, we present a novel approach to clinical NER using BERT-based models distilled from multiple teacher labelers, addressing the computational and scalability challenges associated with deploying large LLMs in clinical settings.  We make three key contributions:

\begin{enumerate}
	\item{We develop teacher labelers combining state-of-the-art LLMs (Gemini and OpenAI models) with medical ontologies (RxNorm and SNOMED) for clinical NER across various note types, validated against expert-labeled datasets.}
	\item{We create and release distilled BERT-based models—approximately 1,000 times smaller than modern LLMs—trained on teacher labels from over 2,000 clinical documents, including oncology progress notes, discharge summaries, radiology reports, and scientific abstracts.}
	\item{We conduct a comprehensive evaluation of our distilled BERT models across five publicly available clinical datasets, including an analysis of model failure modes and an external validation analysis to evaluate the generalizability of our approach across health systems.}
\end{enumerate}

\section*{Methods}

\begin{figure}
\includegraphics[height=3in,width=\linewidth]{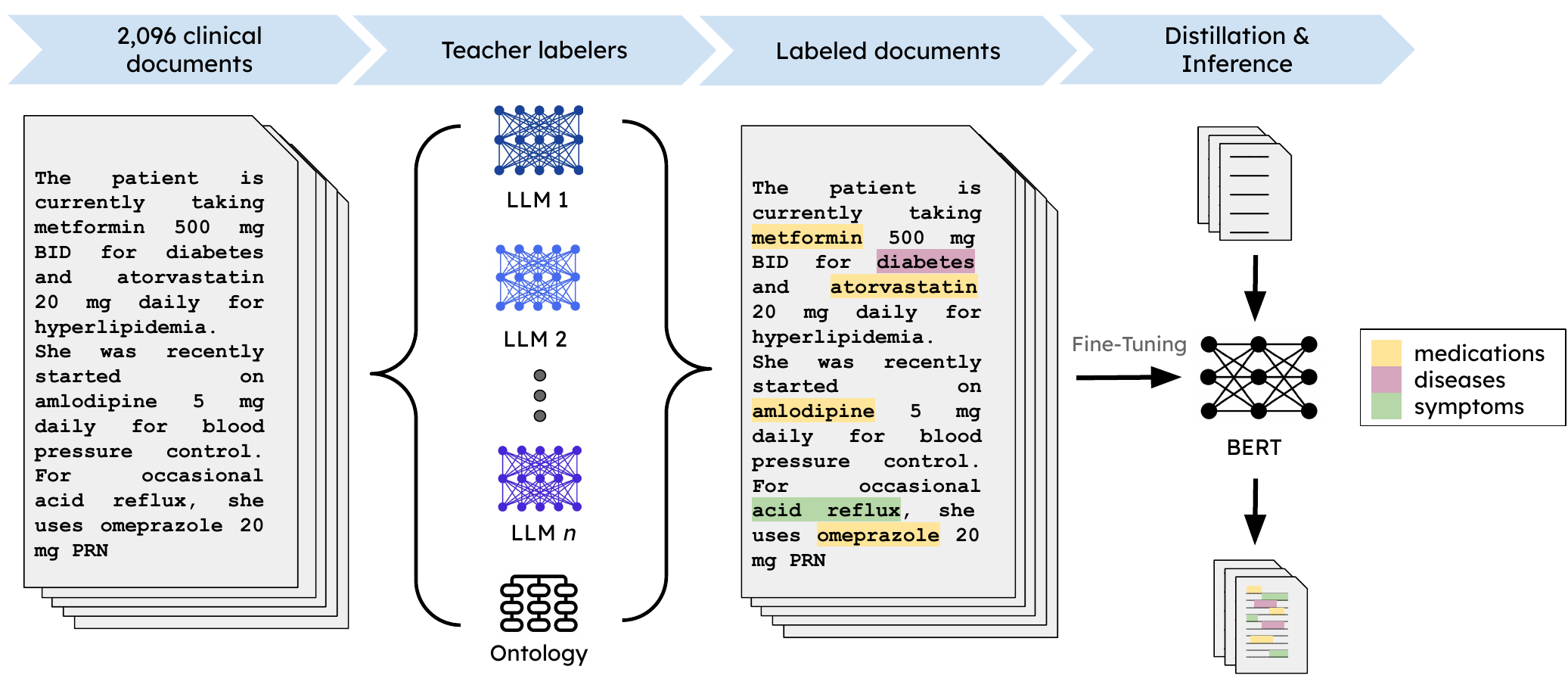}
\caption{Clinical documents were passed to teacher labelers—LLMs and ontologies—for medication, symptom, and disease entity recognition tasks. We selected the optimal combination of teacher labelers based on F1 score for subsequent experiments. BERT models were distilled from the teacher labels via supervised fine-tuning and performance was measured on in-distribution datasets as well as an external validation dataset.}
\end{figure}

This study follows the TRIPOD-LLM\cite{gallifant_tripod-llm_2024} reporting guidelines for the use of LLMs. All experiments were performed with publicly available, de-identified datasets that did not require IRB protocol approval.

\subsection*{NER Tasks and Datasets}
We evaluated our approach on three distinct NER tasks, each utilizing different datasets to ensure comprehensive validation across various clinical contexts:

For the medication extraction task, we used the National NLP Clinical Challenges (n2c2) 2018 Track 2 Medication Extraction dataset.\cite{henry_2018_2020}  This dataset comprises 505 discharge summaries  from MIMIC-III (Medical Information Mart for Intensive Care III)\cite{johnson_mimic-iii_2016}, with expert-annotated medication mentions. Following the n2c2 annotation guidelines, we used 202 notes for testing, 303 notes for training, and randomly sampled 25 notes from the training set for development purposes (Table \ref{tab:DataOverview}). 
The disease extraction task utilized the National Center for Biotechnology Information (NCBI) Disease Corpus.\cite{dogan_ncbi_2014} This corpus contains 793 PubMed abstracts with expert-annotated disease mentions.  We adhered to the official dataset splits for training, development, and testing.
For the symptom extraction task, we used the CORAL dataset\cite{sushil_coral_nodate}, which consists of de-identified progress notes from 40 patients (20 with breast cancer and 20 with pancreatic cancer). These notes, collected at the University of California, San Francisco (UCSF) Information Commons between 2012 and 2022, were de-identified using the Philter tool and annotated at the entity level. We focused on symptoms as they were the most frequent entity type in the dataset. Since CORAL does not provide predefined splits, we randomly selected 5 notes for a development set and 35 for testing, while using the unannotated notes for training through teacher labeling. 

\paragraph{Teacher Labeling Dataset Construction}
Since teacher labeling does not require gold standard annotations, we combined data from all available datasets irrespective of their original annotation status, maximizing data diversity. We leveraged the training splits from our primary datasets (NCBI, n2c2, and CORAL) and augmented them with 1,000 clinical notes sampled from MIMIC-III using a stratified approach to ensure representation across different documentation styles: 250 notes each from progress notes, nursing notes, discharge summaries, and radiology reports. The final teacher labeling dataset used for model fine-tuning consisted of 2,096 documents drawn from NCBI, n2c2, CORAL, and MIMIC-III.

\paragraph{External Validation}
To assess the generalizability of our distilled BERT models, we conducted an external validation study on clinical notes from the MedAlign dataset\cite{fleming_medalign_2023}, a collection of de-identified electronic health records (EHRs) from Stanford Hospital and Lucile Packard Children’s Hospital. From this dataset of 276 longitudinal patient records, we sampled notes across different types to ensure comprehensive evaluation: 250 progress notes, 129 nursing notes, 117 discharge summaries, and 250 procedure notes.
Since MedAlign lacks NER labels, two fourth year medical students (AS and IL) independently  annotated 10 randomly selected notes, with 2 notes doubly annotated to assess inter-rater agreement. Following our model's output format, annotators labeled each token using the Inside-Outside (IO) scheme: “I-MED” for medications, “I-DIS” for disease, “I-SYM” for symptoms, or “O” for all other entities.

\subsection*{Teacher Labeling Pipeline}
\paragraph{LLM-based Labeling}
We evaluated four state-of-the-art LLMs as teacher labelers: GPT-4o (version 2024-08-06)\cite{noauthor_hello_nodate}, GPT-4o-mini (version 2024-07-18)\cite{noauthor_gpt-4o_nodate}, o1-mini (version 2024-09-12)\cite{noauthor_openai_nodate}, and Gemini 1.5 Flash (gemini-1.5-flash-002).\cite{georgiev_gemini_2024}  Each model was prompted to perform the NER tasks and return the extracted entities. All models were executed through HIPAA-compliant API endpoints with standardized parameters (temperature=0.01, top-p=0.9) to ensure consistent outputs. The final optimized prompts are provided in the Supplementary Material.

\paragraph{Ontology-based Labeling}
We leveraged BioPortal\cite{whetzel_bioportal_2011} Annotator API for accessing comprehensive biomedical ontologies: RxNorm\cite{liu_rxnorm_2005} for medication extraction and SNOMED CT\cite{stearns_snomed_2001} for disease and symptom extraction. For each NER task, we mapped the relevant semantic types to their respective tasks, ensuring that only task-relevant entities were extracted. A complete list of the semantic types assigned to each task is provided in the Supplementary Material. 

\paragraph{Optimal Teacher Labeling Regimen}
We hypothesized that different teacher labelers would exhibit varying levels of performance and that an optimal combination of labelers could maximize the F1 score for a given NER task.  Our experiment evaluated all 31 possible subsets of five teacher labelers, comprising four LLM labelers and the ontology labeler. To combine teacher labelers, we took a union of the entities identified by each teacher labeler. For each task and dataset, the combination achieving the highest F1 score on the development set was selected for subsequent experiments.

\subsection*{Model Distillation Implementation}
For each NER task (medication, disease, and symptom extraction), we implemented knowledge distillation using the optimal teacher labeling pipeline to generate training labels. These labels were converted into "Inside-Outside" (IO) format, where words belonging to an entity are labeled as “Inside,” while all other words are labeled as “Outside.” We then fine-tuned separate BERT models for each task using standardized hyperparameters: learning rate=2x10e-5, batch size=8, and weight decay=0.01. All models were trained for 10 epochs on NVIDIA 4xH100 GPUs. The fine-tuned models were used to perform inference on the test sets for each NER task. We report token-level precision, recall, and F1 score, treating the human annotations as the gold standard.
To assess the impact of domain-specific pretraining on downstream performance, we fine-tuned and compared three BERT variants: 

\begin{itemize}
\item{BERT base\cite{devlin_bert_2019}, a general-purpose language model}
\item{BioBERT\cite{lee_biobert_2020}, pretrained on biomedical literature}
\item{BioClinBERT\cite{alsentzer_publicly_2019}, specialized for clinical text}
\end{itemize}

We evaluated the quality of teacher labels by comparing the performance of distilled models fine-tuned on teacher labels with those fine-tuned on human labels. For the medication and disease tasks, we used human-labeled data from the n2c2 and NCBI training sets, respectively. Due to limited human-labeled data in the CORAL dataset, we could not perform this comparison for symptom extraction. We directly evaluated the teacher labeling pipelines by measuring their performance without model distillation.

\subsection*{Error Analysis}
To better characterize model failure modes and estimate the prevalence of labeling errors in the test sets, we conducted an error analysis of the best-performing models for each task. For each false positive (model assigns a non-“O” label, ground truth is “O”) and false negative (model assigns “O”, ground truth is a non-“O” label), model labels were compared to ground truth labels by two annotators (AS and IL, fourth year medical students). Each false positive or false negative was categorized as either “incorrect”, indicating that the model label was truly incorrect; “partially correct”, indicating that the model labels partially overlapped with the ground truth labels for a given entity, but not completely; or “correct”, indicating that the model label was correct and that the ground truth label was incorrect. For each NER task, we randomly annotated 170 false negatives and false positives, including 90 instances that were doubly annotated for inter-rater agreement calculation.

\subsection*{Inference Time and Cost Analysis}
To quantify the practical benefits of deploying smaller models for clinical NER, we compare inference time per note and cost per note between our distilled BERT models and LLM teacher labelers. To calculate cost per note for LLMs, we use input and output API pricing for OpenAI and Google, and we use the tiktoken\cite{noauthor_openaitiktoken_2024} Python library to calculate token counts. We estimate cost per note for BERT models by multiplying inference time by \$28/hour, which is the average cost of a virtual machine with 4xH100 (our compute resources) listed by six cloud vendors\cite{noauthor_cloud_2024} as of September 2024 (Table \ref{tab:GPUCost}).

\section*{Results}

\subsection*{Performance of Teacher Labeler Combinations}
We evaluated all 31 possible combinations of LLM and ontology labelers across our three extraction tasks (Tables \ref{tab:DiseaseComparison} \ref{tab:MedicationComparison} \ref{tab:SymptomComparison}). For symptom extraction, the Gemini 1.5 flash + GPT-4o combination achieved the highest F1 score of 0.801, notably outperforming other combinations including Gemini 1.5 flash + GPT-4o + GPT-4o-mini (F1 = 0.784) and o1-mini + GPT-4o (F1 = 0.778). Interestingly, none of the top-performing combinations for symptom extraction included the ontology-based labeler. 

The medication extraction task showed similar patterns, with Gemini 1.5 flash + GPT-4o achieving the highest F1 score of 0.881. This was followed closely by Gemini-1.5-flash + GPT-4o + GPT-4o-mini (F1 = 0.872) and Gemini-1.5-flash + ontology + GPT-4o (F1 = 0.870). 

For disease extraction, the single o1-mini model achieved the highest F1 score of 0.787, with combinations of o1-mini + ontology (F1 = 0.773) and o1-mini + GPT-4o (F1 = 0.760) performing slightly lower. 

\begin{table}
  \begin{center}
    \begin{tabular}[c]{|l|p{3in}|l|l|l|}
      \hline
      \textbf{Task} & \textbf{Teacher labeler(s)} & \textbf{F1-Score} & \textbf{Precision} & \textbf{Recall} \\
      \hline
      Disease Extraction & o1-mini & 0.787 & 0.724 & 0.862 \\
      & o1-mini + ontology & 0.773 & 0.686 & 0.885 \\
      & o1-mini + GPT-4o & 0.760 & 0.652 & 0.911 \\
      & o1-mini + ontology + GPT-4o & 0.748 & 0.629 & 0.923 \\
      & GPT-4o & 0.748 & 0.717 & 0.781 \\
      \hline
      Medication Extraction & Gemini-1.5-flash + GPT-4o & 0.881 & 0.947 &
      0.824 \\
      & Gemini-1.5-flash + GPT-4o + GPT-4o-mini & 0.872 & 0.896 & 0.849 \\
      & Gemini-1.5-flash + ontology + GPT-4o & 0.870 & 0.865 & 0.876 \\
      & Gemini-1.5-flash + ontology & 0.862 & 0.876 & 0.848 \\
      & Gemini-1.5-flash + GPT-4o-mini & 0.859 & 0.912 & 0.811 \\
      \hline
      Symptom Extraction & Gemini-1.5-flash + GPT-4o & 0.801 & 0.871 & 0.741 \\
      & GPT-4o & 0.787 & 0.900 & 0.700 \\
      & Gemini-1.5-flash + GPT-4o + GPT-4o-mini & 0.784 & 0.810 & 0.759 \\
      & o1-mini + GPT-4o & 0.778 & 0.752 & 0.806 \\
      & o1-mini + Gemini-1.5-flash + GPT-4o & 0.770 & 0.734 & 0.809 \\
      \hline
    \end{tabular}
  \end{center}
  \caption{Top five teacher labeler combinations with highest F1 scores for each NER task.}\label{tab:BestTeachers}
\end{table}

\subsection*{BERT Model Performance}
BioBERT demonstrated superior performance in disease extraction with an F1 score of 0.865, compared to 0.830 for both BaseBERT and BioClinBERT (Table \ref{tab:BERTComparison}).  For the medication extraction task, both BioBERT and BioClinBERT achieved an F1 of 0.89, slightly outperforming BaseBERT (F1 = 0.885).  Symptom extraction proved more challenging across all models, with BioBERT and BioClinBERT achieving F1 scores of 0.34 and BaseBERT reaching 0.33.

\begin{table}[h!]
  \begin{center}
    \begin{tabular}[c]{|l|l|l|l|l|l|l|}
      \hline
      \textbf{Task} & \textbf{Model} & \textbf{F1-Score} & \textbf{NPV} & \textbf{PPV} & \textbf{Sensitivity} & \textbf{Specificity} \\
      \hline
      Disease Extraction & Human + BERT & 0.89 & 0.99 & 0.87 & 0.92 & 0.99 \\
      & Teacher + BERT & 0.84 & 0.99 & 0.78 & 0.90 & 0.98 \\
      & Teacher only & 0.82 & 0.99 & 0.79 & 0.86 & 0.98 \\
      \hline
      Medication Extraction & Human + BERT & 0.91 & 1.00 & 0.89 & 0.93 & 1.00 \\
      & Teacher + BERT & 0.87 & 1.00 & 0.89 & 0.85 & 1.00 \\
      & Teacher only & 0.84 & 0.99 & 0.91 & 0.79 & 1.00 \\
      \hline
      Symptom Extraction & Teacher + BERT & 0.68 & 0.99 & 0.80 & 0.59 & 1.00 \\
      & Teacher only & 0.73 & 0.99 & 0.78 & 0.69 & 1.00 \\
      \hline
    \end{tabular}
  \end{center}
  \caption{Performance of BioBERT models fine-tuned on human labels, BioBERT models distilled from teacher labelers, and the teacher labelers themselves.}\label{tab:BERTvsTeacher}
\end{table}

\subsection*{Comparative Analysis of Label Sources}
Our analysis revealed that models fine-tuned on human labels consistently outperformed those using teacher labels, which in turn exceeded direct teacher labeler performance. For disease extraction, human-labeled models achieved an F1 of 0.89, compared to 0.84 for teacher-labeled models and 0.82 for direct teacher labelers (Table \ref{tab:BERTvsTeacher}. Similarly, in medication extraction, we observed F1 scores of 0.91, 0.87, and 0.84 respectively. For symptom extraction, where human-labeled comparison was not possible, teacher-labeled models achieved an F1 of 0.68, while direct teacher labelers reached 0.73.

\subsection*{Error Analysis}
Across all three tasks, the majority of false positives were due to incorrect ground truth labels. For symptom extraction, 2.20\% of false negatives and 82.05\% of false positives had incorrect ground truth labels (Table \ref{tab:ErrorAnalysis}).  For medication extraction, 21.05\% of false negatives and 62.93\% of false positives had incorrect ground truth labels. For disease extraction, 4.08\% of false negatives and 73.33\% of false positives had incorrect ground truth labels.  More analysis of specific cases for all three tasks are in the Supplementary Material.

\begin{table}
\centering
\begin{tabular}{|l|p{1.5in}|p{1.5in}|p{1.5in}|}
\hline
\textbf{} & \textbf{Symptom Extraction} & \textbf{Medication Extraction} & \textbf{Disease Extraction} \\ \hline
\multicolumn{4}{|c|}{\textbf{\textit{False Negatives}}} \\ \hline
\textbf{N} & 91 & 57 & 49 \\ \hline
\textbf{Correct} & 2.20\% & 21.05\% & 4.08\% \\ \hline
\textbf{Partially Correct} & 16.49\% & 14.04\% & 38.78\% \\ \hline
\textbf{Incorrect} & 81.32\% & 64.91\% & 57.14\% \\ \hline
\textbf{Examples} & 
``She is asymptomatic from \textbf{bone lesions} but we can\ldots'' \newline\newline ``\ldots presented with \textbf{acholic stool} and \textbf{dark urine}'' \newline\newline ``\ldots3 years of \textbf{hot flashes}.'' \newline\newline ``\ldots such as fatigue, neuropathy, \textbf{skin} and nail changes\ldots'' & 
``\ldots{}\textbf{Cardura} 2 q.d\ldots.'' \newline\newline ``DNR / DNI / no \textbf{pressors}\ldots'' \newline\newline ``\textbf{NG} SL PRN'' \newline\newline ``We generally recommend taking an \textbf{over the counter stool softener}\ldots'' & 
``\ldots remarkable propensity to \textbf{bacterial infections}\ldots'' \newline\newline ``Royal National Hospital for \textbf{Rheumatic Diseases} database\ldots'' \newline\newline ``\ldots in three \textbf{prostate cancer} cell lines\ldots'' \newline\newline ``\ldots rescues the \textbf{gastrulation defect}.'' \\ \hline
\multicolumn{4}{|c|}{\textbf{\textit{False Positives}}} \\ \hline
\textbf{N} & 78 & 116 & 120 \\ \hline
\textbf{Correct} & 82.05\% & 62.93\% & 73.33\% \\ \hline
\textbf{Partially Correct} & 15.38\% & 12.93\% & 20.00\% \\ \hline
\textbf{Incorrect} & 2.56\% & 24.14\% & 6.67\% \\ \hline
\textbf{Examples} & 
``\ldots found to have \textbf{sepsis}\ldots'' \newline\newline ``\ldots skin and nail changes, myalgias, alopecia, \textbf{myelosuppression}, nausea\ldots'' & 
``Bilateral injected \textbf{sclera}\ldots'' \newline\newline ``At \textbf{initail} {[}sic{]} deployment the patient\ldots'' & 
``\ldots showed the father to be \textbf{hypohaptoglobinemic}\ldots{}'' \\ \hline
\end{tabular}
  \caption{Error analysis for all NER tasks. A random sample of false positives and false negatives were reviewed and categorized as “Correct” (model label is correct; ground truth label is incorrect), “Partially correct” (model label partially overlaps with ground truth label for the entity), or “Incorrect” (model label is incorrect; ground truth label is correct). Representative examples of false negatives and false positives for each task are shown above. Abbreviations: “NG SL PRN” = nitroglycerin sublingual pro re nata (as needed); “2 q.d.” = two per day.}\label{tab:ErrorAnalysis}
\end{table}

\subsection*{External Validation}
To evaluate the generalizability of our distilled BERT models for clinical NER tasks, we conducted an external validation study on clinical notes sampled from the MedAlign dataset. For disease extraction, the model demonstrated a recall of 89.0\% and a precision of 61.3\%, leading to an F1 of 0.726 (Table \ref{tab:ExternalValidation}).  For medication extraction, the distilled BERT model achieved a recall of 96.4\%, precision of 81.5\%, and F1 of 0.883. Symptom extraction showed the weakest performance, with a recall of 56.0\%, a precision of 92.9\%, and an F1 of 0.699. These results highlight the strong performance of the model for medication and disease extraction tasks, even when applied to an out-of-distribution dataset.

\begin{table}
  \begin{center}
    \begin{tabular}[c]{|l|p{1in}|p{1in}|p{1in}|p{1in}|}
      \hline
      \textbf{Model} & \textbf{Total cost (USD)} & \textbf{Total inference time (s)} & \textbf{Cost per note (USD)} & \textbf{Inference time per note (s)}\\
      \hline
      Distilled BioBERT & 0.02 & 14 & 0.000187 & 0.14 \\
      \hline
      GPT-4o & 1.59 (+7850\%) & 166 (+1086\%) & 0.0159 (+8402\%) & 1.66 (+1086\%) \\
      \hline
      o1-mini & 1.89 (+9350\%) & 58 (+314.3\%) & 0.0189 (+1001\%) & 0.58 (+314.3\%) \\
      \hline
      Gemini 1.5 Flash & 0.05 (+150\%) & 117 (+735.7\%) & 0.000460 (+146.0\%) & 1.17 (+735.7\%) \\
      \hline
    \end{tabular}
  \end{center}
  \caption{On 100 notes sampled from the MedAlign dataset, we report average inference time per note and average cost per note aggregated across medication, disease, and symptom extraction tasks. Parentheses indicate percent difference compared to the Distilled BioBERT model. BioBERT was run on 1xA100 80GB, for which the average cost per hour was estimated at \$4.74 (Table \ref{tab:GPUCost}). For teacher LLMs, we use reported token pricing as of 12/11/2024.}\label{tab:CostComparison}
\end{table}

\subsection*{Inference Time and Cost}
To quantify efficiency gains from our knowledge distillation approach, we compared the inference time per note and cost per note of our distilled BERT models against several teacher labelers, including state-of-the-art LLMs. The distilled BERT model demonstrated superior efficiency, with an average inference time of 0.14 seconds per note and a cost of \$0.000187 per note, calculated based on an estimated \$4.74/hour for a 4xH100 virtual machine (Tables \ref{tab:CostComparison} \ref{tab:GPUCost}). In contrast, teacher LLMs incurred significantly higher inference times and costs: GPT-4o required 1.66 seconds per note and cost \$0.0159 per note; o1-mini model achieved slightly better performance with 0.58 seconds per note and a cost of \$0.0189 per note; and Gemini 1.5 Flash was the cheapest among the teacher labelers, with 1.17 seconds per note and \$0.000460 per note.

\section*{Discussion}

Our study found that distilled BERT models outperformed teacher labelers and approached the performance of BERT models fine tuned on human labels, highlighting the effectiveness of knowledge distillation for clinical NER. In external validation, the distilled BERT models demonstrated strong performance on the medication and disease extraction tasks. Importantly, the distilled BERT models were faster (2x, 4x, 8x faster than GPT-4o, o1-mini, and Gemini Flash respectively) and cheaper (85x, 101x, 2x cheaper than GPT-4o, o1-mini, and Gemini Flash respectively) than their LLM counterparts, making them a practical alternative for real-world clinical applications. Together, these findings highlight the potential of distillation to facilitate efficient and scalable clinical NER while maintaining high performance.

Unlike other studies, which distilled from a single large model, our study assessed 31 different model combinations for different medical NER tasks, and used the best combinations to then distill down to smaller BERT-based models. Additionally, we assessed the effect of including ontology-outputs in the distillation process, finding that their inclusion resulted in poorer performance, due to increased false positives.  We tested these models on discharge summary and medical research publication data, along with an external dataset, demonstrating generalizability.

This study has several limitations. First, the quality of teacher LLMs used to fine-tune the distilled BERT models was often variable, particularly for symptoms. The inconsistency in symptom labeling, particularly between the development and test sets, likely contributed to the lower F1 scores observed for symptom extraction tasks. Second, we focus on only three types of entities; other entity types such as procedures, social determinants of health, diagnosis dates, lab values, and vital signs also need to be extracted for comprehensive clinical information extraction. Third, our approach did not address more complex NER tasks, such as capturing assertion status (e.g., negations or hypothetical statements) or relational extraction tasks (e.g., drug-dosage relationships). Fourth, we did not explore prompt engineering by model and used the same prompts for all LLMs.\cite{jeong_medical_2024}  Finally, the test sets for our three NER tasks have errors. As confirmed by others, they frequently contained labels that were inconsistent with the annotation guidelines of their respective datasets.\cite{gu_distilling_2023}\cite{dogan_ncbi_2014} This inconsistency led to outputs that often did not align with the test set labels, leading to lower performance during evaluation.

An error analysis of the model outputs revealed that human-labeled test sets for all three tasks—medication, disease, and symptom extraction—consistently missed several entities that were correctly identified by the models: 63–82\% of the model's false positives were actually correct, suggesting that the reported precision and F1 scores of our models may be lower bounds.

\section*{Conclusion}

Our work provides a roadmap for leveraging state-of-the-art LLMs to develop efficient, performant, and generalizable clinical NER models through distillation. Ultimately, this study underscores the potential of distilled BERT models as a computationally efficient and scalable alternative to LLMs for clinical NER, paving the way for broader applications in healthcare information extraction.

\section*{References}
\bibliographystyle{naturemag}
\bibliography{distill2024}

\begin{thebibliography}{10}
\expandafter\ifx\csname url\endcsname\relax
  \def\url#1{\texttt{#1}}\fi
\expandafter\ifx\csname urlprefix\endcsname\relax\def\urlprefix{URL }\fi
\providecommand{\bibinfo}[2]{#2}
\providecommand{\eprint}[2][]{\url{#2}}

\bibitem{ross_big_2014}
\bibinfo{author}{Ross, M.~K.}, \bibinfo{author}{Wei, W.} \& \bibinfo{author}{Ohno-Machado, L.}
\newblock \bibinfo{title}{"{Big} data" and the electronic health record}.
\newblock \emph{\bibinfo{journal}{Yearbook of Medical Informatics}} \textbf{\bibinfo{volume}{9}}, \bibinfo{pages}{97--104} (\bibinfo{year}{2014}).

\bibitem{wornow_zero-shot_2024}
\bibinfo{author}{Wornow, M.} \emph{et~al.}
\newblock \bibinfo{title}{Zero-{Shot} {Clinical} {Trial} {Patient} {Matching} with {LLMs}} (\bibinfo{year}{2024}).
\newblock \urlprefix\url{http://arxiv.org/abs/2402.05125}.
\newblock \bibinfo{note}{ArXiv:2402.05125 [cs]}.

\bibitem{callahan_research_2020}
\bibinfo{author}{Callahan, A.}, \bibinfo{author}{Shah, N.~H.} \& \bibinfo{author}{Chen, J.~H.}
\newblock \bibinfo{title}{Research and {Reporting} {Considerations} for {Observational} {Studies} {Using} {Electronic} {Health} {Record} {Data}}.
\newblock \emph{\bibinfo{journal}{Annals of Internal Medicine}} \textbf{\bibinfo{volume}{172}}, \bibinfo{pages}{S79--S84} (\bibinfo{year}{2020}).

\bibitem{lamurias_lasigebiotm_2019}
\bibinfo{author}{Lamurias, A.} \& \bibinfo{author}{Couto, F.~M.}
\newblock \bibinfo{title}{{LasigeBioTM} at {MEDIQA} 2019: {Biomedical} {Question} {Answering} using {Bidirectional} {Transformers} and {Named} {Entity} {Recognition}}.
\newblock In \emph{\bibinfo{booktitle}{Proceedings of the 18th {BioNLP} {Workshop} and {Shared} {Task}}}, \bibinfo{pages}{523--527} (\bibinfo{publisher}{Association for Computational Linguistics}, \bibinfo{address}{Florence, Italy}, \bibinfo{year}{2019}).
\newblock \urlprefix\url{https://www.aclweb.org/anthology/W19-5057}.

\bibitem{zweigenbaum_frontiers_2007}
\bibinfo{author}{Zweigenbaum, P.}, \bibinfo{author}{Demner-Fushman, D.}, \bibinfo{author}{Yu, H.} \& \bibinfo{author}{Cohen, K.~B.}
\newblock \bibinfo{title}{Frontiers of biomedical text mining: current progress}.
\newblock \emph{\bibinfo{journal}{Briefings in bioinformatics}} \textbf{\bibinfo{volume}{8}}, \bibinfo{pages}{358--375} (\bibinfo{year}{2007}).
\newblock \urlprefix\url{https://www.ncbi.nlm.nih.gov/pmc/articles/PMC2516302/}.

\bibitem{wang_clinical_2018}
\bibinfo{author}{Wang, Y.} \emph{et~al.}
\newblock \bibinfo{title}{Clinical information extraction applications: {A} literature review}.
\newblock \emph{\bibinfo{journal}{Journal of Biomedical Informatics}} \textbf{\bibinfo{volume}{77}}, \bibinfo{pages}{34--49} (\bibinfo{year}{2018}).

\bibitem{lample_neural_2016}
\bibinfo{author}{Lample, G.}, \bibinfo{author}{Ballesteros, M.}, \bibinfo{author}{Subramanian, S.}, \bibinfo{author}{Kawakami, K.} \& \bibinfo{author}{Dyer, C.}
\newblock \bibinfo{title}{Neural {Architectures} for {Named} {Entity} {Recognition}} (\bibinfo{year}{2016}).
\newblock \urlprefix\url{http://arxiv.org/abs/1603.01360}.
\newblock \bibinfo{note}{ArXiv:1603.01360 [cs]}.

\bibitem{bodenreider_unified_2004}
\bibinfo{author}{Bodenreider, O.}
\newblock \bibinfo{title}{The {Unified} {Medical} {Language} {System} ({UMLS}): integrating biomedical terminology}.
\newblock \emph{\bibinfo{journal}{Nucleic Acids Research}} \textbf{\bibinfo{volume}{32}}, \bibinfo{pages}{D267--D270} (\bibinfo{year}{2004}).
\newblock \urlprefix\url{https://www.ncbi.nlm.nih.gov/pmc/articles/PMC308795/}.

\bibitem{liao_development_2015}
\bibinfo{author}{Liao, K.~P.} \emph{et~al.}
\newblock \bibinfo{title}{Development of phenotype algorithms using electronic medical records and incorporating natural language processing}.
\newblock \emph{\bibinfo{journal}{The BMJ}} \textbf{\bibinfo{volume}{350}}, \bibinfo{pages}{h1885} (\bibinfo{year}{2015}).
\newblock \urlprefix\url{https://www.ncbi.nlm.nih.gov/pmc/articles/PMC4707569/}.

\bibitem{campillos-llanos_clinical_2021}
\bibinfo{author}{Campillos-Llanos, L.}, \bibinfo{author}{Valverde-Mateos, A.}, \bibinfo{author}{Capllonch-Carrión, A.} \& \bibinfo{author}{Moreno-Sandoval, A.}
\newblock \bibinfo{title}{A clinical trials corpus annotated with {UMLS} entities to enhance the access to evidence-based medicine}.
\newblock \emph{\bibinfo{journal}{BMC Medical Informatics and Decision Making}} \textbf{\bibinfo{volume}{21}}, \bibinfo{pages}{69} (\bibinfo{year}{2021}).
\newblock \urlprefix\url{https://doi.org/10.1186/s12911-021-01395-z}.

\bibitem{devlin_bert_2019}
\bibinfo{author}{Devlin, J.}, \bibinfo{author}{Chang, M.-W.}, \bibinfo{author}{Lee, K.} \& \bibinfo{author}{Toutanova, K.}
\newblock \bibinfo{title}{{BERT}: {Pre}-training of {Deep} {Bidirectional} {Transformers} for {Language} {Understanding}}.
\newblock In \bibinfo{editor}{Burstein, J.}, \bibinfo{editor}{Doran, C.} \& \bibinfo{editor}{Solorio, T.} (eds.) \emph{\bibinfo{booktitle}{Proceedings of the 2019 {Conference} of the {North} {American} {Chapter} of the {Association} for {Computational} {Linguistics}: {Human} {Language} {Technologies}, {Volume} 1 ({Long} and {Short} {Papers})}}, \bibinfo{pages}{4171--4186} (\bibinfo{publisher}{Association for Computational Linguistics}, \bibinfo{address}{Minneapolis, Minnesota}, \bibinfo{year}{2019}).
\newblock \urlprefix\url{https://aclanthology.org/N19-1423.pdf}.

\bibitem{fries_ontology-driven_2021}
\bibinfo{author}{Fries, J.~A.} \emph{et~al.}
\newblock \bibinfo{title}{Ontology-driven weak supervision for clinical entity classification in electronic health records}.
\newblock \emph{\bibinfo{journal}{Nature Communications}} \textbf{\bibinfo{volume}{12}}, \bibinfo{pages}{2017} (\bibinfo{year}{2021}).
\newblock \urlprefix\url{https://www.nature.com/articles/s41467-021-22328-4}.
\newblock \bibinfo{note}{Publisher: Nature Publishing Group}.

\bibitem{lee_biobert_2020}
\bibinfo{author}{Lee, J.} \emph{et~al.}
\newblock \bibinfo{title}{{BioBERT}: a pre-trained biomedical language representation model for biomedical text mining}.
\newblock \emph{\bibinfo{journal}{Bioinformatics}} \textbf{\bibinfo{volume}{36}}, \bibinfo{pages}{1234--1240} (\bibinfo{year}{2020}).
\newblock \urlprefix\url{https://doi.org/10.1093/bioinformatics/btz682}.

\bibitem{huang_clinicalbert_2020}
\bibinfo{author}{Huang, K.}, \bibinfo{author}{Altosaar, J.} \& \bibinfo{author}{Ranganath, R.}
\newblock \bibinfo{title}{{ClinicalBERT}: {Modeling} {Clinical} {Notes} and {Predicting} {Hospital} {Readmission}} (\bibinfo{year}{2020}).
\newblock \urlprefix\url{http://arxiv.org/abs/1904.05342}.
\newblock \bibinfo{note}{ArXiv:1904.05342 [cs]}.

\bibitem{chaves_rales_2023}
\bibinfo{author}{Chaves, J. M.~Z.} \emph{et~al.}
\newblock \bibinfo{title}{{RaLEs}: a {Benchmark} for {Radiology} {Language} {Evaluations}}.
\newblock In \emph{\bibinfo{booktitle}{Thirty-seventh Conference on Neural Information Processing Systems Datasets and Benchmarks Track}} (\bibinfo{year}{2023}).
\newblock \urlprefix\url{https://openreview.net/forum?id=PWLGrvoqiR}.

\bibitem{monajatipoor_llms_2024}
\bibinfo{author}{Monajatipoor, M.} \emph{et~al.}
\newblock \bibinfo{title}{{LLMs} in {Biomedicine}: {A} study on clinical {Named} {Entity} {Recognition}} (\bibinfo{year}{2024}).
\newblock \urlprefix\url{http://arxiv.org/abs/2404.07376}.
\newblock \bibinfo{note}{ArXiv:2404.07376 [cs]}.

\bibitem{noauthor_pricing_nodate}
\bibinfo{title}{Pricing}.
\newblock \urlprefix\url{https://openai.com/api/pricing/}.

\bibitem{hinton_distilling_2015}
\bibinfo{author}{Hinton, G.}, \bibinfo{author}{Vinyals, O.} \& \bibinfo{author}{Dean, J.}
\newblock \bibinfo{title}{Distilling the {Knowledge} in a {Neural} {Network}} (\bibinfo{year}{2015}).
\newblock \urlprefix\url{http://arxiv.org/abs/1503.02531}.
\newblock \bibinfo{note}{ArXiv:1503.02531 [stat]}.

\bibitem{zhou_universalner_2023}
\bibinfo{author}{Zhou, W.}, \bibinfo{author}{Zhang, S.}, \bibinfo{author}{Gu, Y.}, \bibinfo{author}{Chen, M.} \& \bibinfo{author}{Poon, H.}
\newblock \bibinfo{title}{Universal{NER}: Targeted distillation from large language models for open named entity recognition}.
\newblock In \emph{\bibinfo{booktitle}{The Twelfth International Conference on Learning Representations}} (\bibinfo{year}{2024}).
\newblock \urlprefix\url{https://openreview.net/forum?id=r65xfUb76p}.

\bibitem{rhouma_leveraging_2024}
\bibinfo{author}{Rhouma, R.} \emph{et~al.}
\newblock \bibinfo{title}{Leveraging mobile {NER} for real-time capture of symptoms, diagnoses, and treatments from clinical dialogues}.
\newblock \emph{\bibinfo{journal}{Informatics in Medicine Unlocked}} \textbf{\bibinfo{volume}{48}}, \bibinfo{pages}{101519} (\bibinfo{year}{2024}).
\newblock \urlprefix\url{https://www.sciencedirect.com/science/article/pii/S2352914824000753}.

\bibitem{gu_distilling_2023}
\bibinfo{author}{Gu, Y.} \emph{et~al.}
\newblock \bibinfo{title}{Distilling {Large} {Language} {Models} for {Biomedical} {Knowledge} {Extraction}: {A} {Case} {Study} on {Adverse} {Drug} {Events}} (\bibinfo{year}{2023}).
\newblock \urlprefix\url{http://arxiv.org/abs/2307.06439}.
\newblock \bibinfo{note}{ArXiv:2307.06439 [cs]}.

\bibitem{gallifant_tripod-llm_2024}
\bibinfo{author}{Gallifant, J.} \emph{et~al.}
\newblock \bibinfo{title}{The {TRIPOD}-{LLM} {Statement}: {A} {Targeted} {Guideline} {For} {Reporting} {Large} {Language} {Models} {Use}}.
\newblock \emph{\bibinfo{journal}{medRxiv: The Preprint Server for Health Sciences}} \bibinfo{pages}{2024.07.24.24310930} (\bibinfo{year}{2024}).

\bibitem{henry_2018_2020}
\bibinfo{author}{Henry, S.}, \bibinfo{author}{Buchan, K.}, \bibinfo{author}{Filannino, M.}, \bibinfo{author}{Stubbs, A.} \& \bibinfo{author}{Uzuner, O.}
\newblock \bibinfo{title}{2018 n2c2 shared task on adverse drug events and medication extraction in electronic health records}.
\newblock \emph{\bibinfo{journal}{Journal of the American Medical Informatics Association: JAMIA}} \textbf{\bibinfo{volume}{27}}, \bibinfo{pages}{3--12} (\bibinfo{year}{2020}).

\bibitem{johnson_mimic-iii_2016}
\bibinfo{author}{Johnson, A.~E.} \emph{et~al.}
\newblock \bibinfo{title}{{MIMIC}-{III}, a freely accessible critical care database}.
\newblock \emph{\bibinfo{journal}{Scientific Data}} \textbf{\bibinfo{volume}{3}}, \bibinfo{pages}{160035} (\bibinfo{year}{2016}).
\newblock \urlprefix\url{https://www.nature.com/articles/sdata201635}.

\bibitem{dogan_ncbi_2014}
\bibinfo{author}{Doğan, R.~I.}, \bibinfo{author}{Leaman, R.} \& \bibinfo{author}{Lu, Z.}
\newblock \bibinfo{title}{{NCBI} disease corpus: a resource for disease name recognition and concept normalization}.
\newblock \emph{\bibinfo{journal}{Journal of Biomedical Informatics}} \textbf{\bibinfo{volume}{47}}, \bibinfo{pages}{1--10} (\bibinfo{year}{2014}).

\bibitem{sushil_coral_nodate}
\bibinfo{author}{Sushil, M.} \emph{et~al.}
\newblock \bibinfo{title}{{CORAL}: expert-{Curated} medical {Oncology} {Reports} to {Advance} {Language} model inference}.
\newblock \urlprefix\url{https://physionet.org/content/curated-oncology-reports/1.0/}.

\bibitem{fleming_medalign_2023}
\bibinfo{author}{Fleming, S.~L.} \emph{et~al.}
\newblock \bibinfo{title}{{MedAlign}: {A} {Clinician}-{Generated} {Dataset} for {Instruction} {Following} with {Electronic} {Medical} {Records}} (\bibinfo{year}{2023}).
\newblock \urlprefix\url{http://arxiv.org/abs/2308.14089}.
\newblock \bibinfo{note}{ArXiv:2308.14089 [cs]}.

\bibitem{noauthor_hello_nodate}
\bibinfo{title}{Hello {GPT}-4o}.
\newblock \urlprefix\url{https://openai.com/index/hello-gpt-4o/}.

\bibitem{noauthor_gpt-4o_nodate}
\bibinfo{title}{{GPT}-4o mini: advancing cost-efficient intelligence}.
\newblock \urlprefix\url{https://openai.com/index/gpt-4o-mini-advancing-cost-efficient-intelligence/}.

\bibitem{noauthor_openai_nodate}
\bibinfo{title}{{OpenAI} o1-mini}.
\newblock \urlprefix\url{https://openai.com/index/openai-o1-mini-advancing-cost-efficient-reasoning/}.

\bibitem{georgiev_gemini_2024}
\bibinfo{author}{Georgiev, P.} \emph{et~al.}
\newblock \bibinfo{title}{Gemini 1.5: {Unlocking} multimodal understanding across millions of tokens of context} (\bibinfo{year}{2024}).
\newblock \urlprefix\url{http://arxiv.org/abs/2403.05530}.
\newblock \bibinfo{note}{ArXiv:2403.05530 [cs]}.

\bibitem{whetzel_bioportal_2011}
\bibinfo{author}{Whetzel, P.~L.} \emph{et~al.}
\newblock \bibinfo{title}{{BioPortal}: enhanced functionality via new {Web} services from the {National} {Center} for {Biomedical} {Ontology} to access and use ontologies in software applications}.
\newblock \emph{\bibinfo{journal}{Nucleic Acids Research}} \textbf{\bibinfo{volume}{39}}, \bibinfo{pages}{W541--545} (\bibinfo{year}{2011}).

\bibitem{liu_rxnorm_2005}
\bibinfo{author}{Liu, S.}, \bibinfo{author}{{Wei Ma}}, \bibinfo{author}{Moore, R.}, \bibinfo{author}{Ganesan, V.} \& \bibinfo{author}{Nelson, S.}
\newblock \bibinfo{title}{{RxNorm}: prescription for electronic drug information exchange}.
\newblock \emph{\bibinfo{journal}{IT Professional}} \textbf{\bibinfo{volume}{7}}, \bibinfo{pages}{17--23} (\bibinfo{year}{2005}).
\newblock \urlprefix\url{http://ieeexplore.ieee.org/document/1516084/}.

\bibitem{stearns_snomed_2001}
\bibinfo{author}{Stearns, M.~Q.}, \bibinfo{author}{Price, C.}, \bibinfo{author}{Spackman, K.~A.} \& \bibinfo{author}{Wang, A.~Y.}
\newblock \bibinfo{title}{{SNOMED} clinical terms: overview of the development process and project status.}
\newblock \emph{\bibinfo{journal}{Proceedings of the AMIA Symposium}} \bibinfo{pages}{662--666} (\bibinfo{year}{2001}).
\newblock \urlprefix\url{https://www.ncbi.nlm.nih.gov/pmc/articles/PMC2243297/}.

\bibitem{alsentzer_publicly_2019}
\bibinfo{author}{Alsentzer, E.} \emph{et~al.}
\newblock \bibinfo{title}{Publicly {Available} {Clinical} {BERT} {Embeddings}}.
\newblock In \bibinfo{editor}{Rumshisky, A.}, \bibinfo{editor}{Roberts, K.}, \bibinfo{editor}{Bethard, S.} \& \bibinfo{editor}{Naumann, T.} (eds.) \emph{\bibinfo{booktitle}{Proceedings of the 2nd {Clinical} {Natural} {Language} {Processing} {Workshop}}}, \bibinfo{pages}{72--78} (\bibinfo{publisher}{Association for Computational Linguistics}, \bibinfo{address}{Minneapolis, Minnesota, USA}, \bibinfo{year}{2019}).
\newblock \urlprefix\url{https://aclanthology.org/W19-1909}.

\bibitem{noauthor_openaitiktoken_2024}
\bibinfo{title}{openai/tiktoken} (\bibinfo{year}{2024}).
\newblock \urlprefix\url{https://github.com/openai/tiktoken}.
\newblock \bibinfo{note}{Original-date: 2022-12-01T23:22:11Z}.

\bibitem{noauthor_cloud_2024}
\bibinfo{title}{Cloud {GPU} {Pricing} {Comparison} in 2024} (\bibinfo{year}{2024}).
\newblock \urlprefix\url{https://datacrunch.io/blog/cloud-gpu-pricing-comparison-in-2024}.
\newblock \bibinfo{note}{Section: GPUs}.

\bibitem{jeong_medical_2024}
\bibinfo{author}{Jeong, D.~P.}, \bibinfo{author}{Garg, S.}, \bibinfo{author}{Lipton, Z.~C.} \& \bibinfo{author}{Oberst, M.}
\newblock \bibinfo{title}{Medical {Adaptation} of {Large} {Language} and {Vision}-{Language} {Models}: {Are} {We} {Making} {Progress}?} (\bibinfo{year}{2024}).
\newblock \urlprefix\url{http://arxiv.org/abs/2411.04118}.
\newblock \bibinfo{note}{ArXiv:2411.04118 [cs]}.

\bibitem{noauthor_tui_nodate}
\bibinfo{title}{{TUI} {Semantic} {Type} {List}}.
\newblock \urlprefix\url{https://lhncbc.nlm.nih.gov/ii/tools/MetaMap/Docs/SemanticTypes_2018AB.txt}.

\end{thebibliography}

\section*{Author Contributions}
The study was conceptualized by KSV and AS. Coding and data analysis were performed by KSV, AS, and AG, while data annotation was carried out by AS and IL. All authors contributed to the interpretation of the data and the writing of the manuscript. Supervision was provided by NHS.

\section*{Funding}
No funding was obtained for this study.

\section*{Competing Interests}

AS is a paid advisor to Daybreak Health, holds stock options in Cerebral and Daybreak Health, and holds stock in Roche (RHHVF). NHS reported being a cofounder of Prealize Health (a predictive analytics company), Atropos Health (an on-demand evidence generation company) and serving on the Board of the Coalition for Healthcare AI (CHAI), a consensus-building organization providing guidelines for the responsible use of artificial intelligence in health care. NHS serves as a scientific advisor to Opala, Curai Health, Arsenal Capital and JnJ Innovative Medicines.

\section*{Code Availability}

The code for our experiments can be found at \href{https://github.com/som-shahlab/distill-ner}{\texttt{https://github.com/som-shahlab/distill-ner}}.

\newpage

\section*{Supplementary Material}
\subsection*{Error Analysis of Selected Entities}

\paragraph{Disease Extraction} False negatives like “remarkable propensity to bacterial infections” and “...rescues the gastrulation defect,” suggest difficulty in identifying generic disease categories. False positives, such as “hypohaptoglobinemic,” indicate challenges in distinguishing disease entities from lab findings.

\paragraph{Medication Extraction} False negatives like “Cardura 2 q.d.” may reflect difficulty in identifying uncommon medication names, while missed phrases like “NG SL PRN” suggest difficulty in identifying abbreviations. Descriptions of medication classes or types were also missed, such as “over-the-counter stool softener” or “pressors”. False positives such as “Bilateral injected sclera” reflect confusion of the anatomic sclera with an injected drug, while “At initail deployment” may indicate confusing a misspelling with a drug name. 

\paragraph{Symptom extraction}  False negatives like “acholic stool”, “dark urine”, and “hot flashes” suggest challenges in capturing less common symptoms, while partial mentions like “skin and nail changes” indicate difficulty in handling symptoms embedded within lists. False positives such as “sepsis” and “myelosuppression” reflect confusion between symptoms and diagnoses or treatment side effects.

\subsection*{Semantic Type Unique Identifiers}

The following semantic type unique identifiers (TUIs) were used for the ontology teacher labelers.\cite{noauthor_tui_nodate}

\begin{table}[H]
    \centering
    \begin{tabular}{|l|p{3in}|l|}
        \hline
        \textbf{Entity Type} & \textbf{TUIs} & \textbf{Ontology} \\
        \hline
        Medications & 
        \begin{itemize}
            \item T195 - Antibiotic (antb)
            \item T123 - Biologically Active Substance (bacs)
            \item T200 - Clinical Drug (clnd)
            \item T125 - Hormone (horm)
            \item T121 - Pharmacologic Substance (phsu)
        \end{itemize}
        & RxNorm \\
        \hline
        Diseases & 
        \begin{itemize}
            \item T020 - Acquired Abnormality (acab)
            \item T190 - Anatomical Abnormality (anab)
            \item T019 - Congenital Abnormality (cgab)
            \item T047 - Disease or Syndrome (dsyn)
            \item T050 - Experimental Model of Disease (emod)
            \item T037 - Injury or Poisoning (inpo)
            \item T191 - Neoplastic Process (neop)
            \item T046 - Pathologic Function (patf)
        \end{itemize}
        & SNOMED CT \\
        \hline
        Symptoms & 
        \begin{itemize}
            \item T184 - Sign or Symptom (sosy)
        \end{itemize}
        & SNOMED CT \\
        \hline
    \end{tabular}
    \caption{TUIs used for each Semantic Type \& Ontology}
    \label{tab:entity_types}
\end{table}

\subsection*{Data Tables}

\begin{table}[H]
  \begin{center}
    \begin{tabular}[c]{|l|l|p{3in}|}
      \hline
      \textbf{Vendor} & \textbf{1xA100 80GB cost per hour (USD)} & \textbf{Notes} \\
      \hline
      Google Cloud & \$5.58 & Available in europe-west4 region. \\
      Amazon AWS & \$8.19 & Based on pricing for A100 40GB x8 (\$32.77/hr)
      available in us-east-1 region. \\
      Microsoft Azure & \$3.67 & Available in East US region. \\
      OVHcloud & \$3.07 & \\
      Paperspace & \$3.18 & Based on pricing for A100 80GB x8 (\$25.44/hr). \\
      \hline
      \textbf{Average} & \textbf{\$4.74} & \\
      \hline
    \end{tabular}
  \end{center}
  \caption{GPU cost estimates across cloud vendors, published September 20, 2024 by Data Crunch.}\label{tab:GPUCost}
\end{table}

\begin{table}[H]
  \begin{center}
    \begin{tabular}[c]{|p{3in}|l|l|l|}
      \hline
      \textbf{Model} & \textbf{F1-Score} & \textbf{Precision} & \textbf{Recall} \\
      \hline
      o1-mini & 0.787 & 0.724 & 0.862 \\
      \hline
      o1-mini + ontology & 0.773 & 0.686 & 0.885 \\
      \hline
      o1-mini + gpt-4o & 0.760 & 0.652 & 0.911 \\
      \hline
      o1-mini + ontology + gpt-4o & 0.748 & 0.629 & 0.923 \\
      \hline
      gpt-4o & 0.748 & 0.717 & 0.781 \\
      \hline
      ontology + gpt-4o & 0.738 & 0.682 & 0.803 \\
      \hline
      o1-mini + gemini-1.5-flash & 0.706 & 0.583 & 0.894 \\
      \hline
      o1-mini + gemini-1.5-flash + ontology & 0.693 & 0.561 & 0.905 \\
      \hline
      o1-mini + gemini-1.5-flash + gpt-4o & 0.686 & 0.547 & 0.918 \\
      \hline
      o1-mini + gemini-1.5-flash + ontology + gpt-4o & 0.677 & 0.532 & 0.928 \\
      \hline
      gemini-1.5-flash + gpt-4o & 0.664 & 0.562 & 0.811 \\
      \hline
      gemini-1.5-flash + ontology + gpt-4o & 0.658 & 0.546 & 0.830 \\
      \hline
      gemini-1.5-flash + ontology & 0.638 & 0.580 & 0.707 \\
      \hline
      gemini-1.5-flash & 0.634 & 0.605 & 0.665 \\
      \hline
      o1-mini + gpt-4o-mini & 0.632 & 0.487 & 0.901 \\
      \hline
      o1-mini + gpt-4o + gpt-4o-mini & 0.622 & 0.469 & 0.924 \\
      \hline
      o1-mini + ontology + gpt-4o-mini & 0.622 & 0.473 & 0.909 \\
      \hline
      o1-mini + ontology + gpt-4o + gpt-4o-mini & 0.613 & 0.458 & 0.927 \\
      \hline
      o1-mini + gemini-1.5-flash + gpt-4o-mini & 0.613 & 0.461 & 0.914 \\
      \hline
      o1-mini + gemini-1.5-flash + gpt-4o + gpt-4o-mini & 0.603 & 0.446 & 0.928 \\
      \hline
      o1-mini + gemini-1.5-flash + ontology + gpt-4o-mini & 0.601 & 0.447 & 0.917 \\
      \hline
      gpt-4o + gpt-4o-mini & 0.598 & 0.466 & 0.831 \\
      \hline
      o1-mini + gemini-1.5-flash + ontology + gpt-4o + gpt-4o-mini & 0.594 & 0.436 & 0.931 \\
      \hline
      ontology + gpt-4o + gpt-4o-mini & 0.590 & 0.455 & 0.839 \\
      \hline
      gemini-1.5-flash + gpt-4o + gpt-4o-mini & 0.581 & 0.443 & 0.846 \\
      \hline
      gemini-1.5-flash + ontology + gpt-4o + gpt-4o-mini & 0.575 & 0.433 & 0.854 \\
      \hline
      gemini-1.5-flash + gpt-4o-mini & 0.566 & 0.450 & 0.762 \\
      \hline
      ontology + gpt-4o-mini & 0.564 & 0.458 & 0.732 \\
      \hline
      gemini-1.5-flash + ontology + gpt-4o-mini & 0.561 & 0.438 & 0.779 \\
      \hline
      gpt-4o-mini & 0.557 & 0.466 & 0.694 \\
      \hline
      ontology & 0.499 & 0.761 & 0.371 \\
      \hline
    \end{tabular}
  \end{center}
  \caption{Performance metrics (F1-score, precision, and recall) for all model configurations evaluated on the Disease Extraction task, sorted by F1-score from highest to lowest.}\label{tab:DiseaseComparison}
\end{table}

\begin{table}[H]
  \begin{center}
      \begin{tabular}[c]{|p{3in}|l|l|l|}
            \hline
            \textbf{Model} & \textbf{F1-Score} & \textbf{Precision} & \textbf{Recall} \\
            \hline
            gemini-1.5-flash + gpt-4o & 0.881 & 0.947 & 0.824 \\
            \hline
            gemini-1.5-flash + gpt-4o + gpt-4o-mini & 0.872 & 0.896 & 0.849 \\
            \hline
            gemini-1.5-flash + ontology + gpt-4o & 0.870 & 0.865 & 0.876 \\
            \hline
            gemini-1.5-flash + ontology & 0.862 & 0.876 & 0.848 \\
            \hline
            gemini-1.5-flash + gpt-4o-mini & 0.859 & 0.912 & 0.811 \\
            \hline
            ontology + gpt-4o & 0.857 & 0.869 & 0.845 \\
            \hline
            gemini-1.5-flash + ontology + gpt-4o + gpt-4o-mini & 0.856 & 0.824 & 0.889 \\
            \hline
            gemini-1.5-flash + ontology + gpt-4o-mini & 0.854 & 0.835 & 0.874 \\
            \hline
            gemini-1.5-flash & 0.852 & 0.969 & 0.760 \\
            \hline
            ontology + gpt-4o + gpt-4o-mini & 0.848 & 0.826 & 0.872 \\
            \hline
            gpt-4o + gpt-4o-mini & 0.848 & 0.897 & 0.804 \\
            \hline
            gpt-4o & 0.838 & 0.955 & 0.747 \\
            \hline
            ontology + gpt-4o-mini & 0.826 & 0.833 & 0.819 \\
            \hline
            ontology & 0.766 & 0.868 & 0.684 \\
            \hline
            gpt-4o-mini & 0.762 & 0.906 & 0.657 \\
            \hline
            o1-mini + gemini-1.5-flash + gpt-4o & 0.707 & 0.597 & 0.867 \\
            \hline
            o1-mini + gemini-1.5-flash + gpt-4o + gpt-4o-mini & 0.702 & 0.582 & 0.883 \\
            \hline
            o1-mini + gemini-1.5-flash + ontology + gpt-4o & 0.699 & 0.571 & 0.900 \\
            \hline
            o1-mini + gemini-1.5-flash & 0.696 & 0.595 & 0.839 \\
            \hline
            o1-mini + gemini-1.5-flash + ontology & 0.696 & 0.572 & 0.890 \\
            \hline
            o1-mini + gemini-1.5-flash + gpt-4o-mini & 0.694 & 0.581 & 0.862 \\
            \hline
            o1-mini + ontology + gpt-4o & 0.693 & 0.570 & 0.885 \\
            \hline
            o1-mini + gemini-1.5-flash + ontology + gpt-4o + gpt-4o-mini & 0.691 & 0.557 & 0.910 \\
            \hline
            o1-mini + gemini-1.5-flash + ontology + gpt-4o-mini & 0.691 & 0.558 & 0.905 \\
            \hline
            o1-mini + gpt-4o + gpt-4o-mini & 0.690 & 0.577 & 0.858 \\
            \hline
            o1-mini + gpt-4o & 0.688 & 0.589 & 0.827 \\
            \hline
            o1-mini + ontology + gpt-4o + gpt-4o-mini & 0.688 & 0.556 & 0.901 \\
            \hline
            o1-mini + ontology + gpt-4o-mini & 0.685 & 0.557 & 0.891 \\
            \hline
            o1-mini + ontology & 0.675 & 0.563 & 0.844 \\
            \hline
            o1-mini + gpt-4o-mini & 0.669 & 0.569 & 0.811 \\
            \hline
            o1-mini & 0.611 & 0.551 & 0.685 \\
            \hline
    \end{tabular}
  \end{center}
  \caption{Performance metrics (F1-score, precision, and recall) for all model configurations evaluated on the Medication Extraction task, sorted by F1-score from highest to lowest.}\label{tab:MedicationComparison}
\end{table}

\begin{table}[H]
  \begin{center}
      \begin{tabular}[c]{|p{3in}|l|l|l|}
            \hline
            \textbf{Model} & \textbf{F1-Score} & \textbf{Precision} & \textbf{Recall} \\
            \hline
            gemini-1.5-flash + gpt-4o & 0.801 & 0.871 & 0.741 \\
            \hline
            gpt-4o & 0.787 & 0.900 & 0.700 \\
            \hline
            gemini-1.5-flash + gpt-4o + gpt-4o-mini & 0.784 & 0.810 & 0.759 \\
            \hline
            o1-mini + gpt-4o & 0.778 & 0.752 & 0.806 \\
            \hline
            o1-mini + gemini-1.5-flash + gpt-4o & 0.770 & 0.734 & 0.809 \\
            \hline
            gemini-1.5-flash + ontology + gpt-4o & 0.768 & 0.787 & 0.750 \\
            \hline
            gpt-4o + gpt-4o-mini & 0.767 & 0.819 & 0.722 \\
            \hline
            o1-mini + gpt-4o + gpt-4o-mini & 0.764 & 0.716 & 0.819 \\
            \hline
            o1-mini + gpt-4o-mini & 0.763 & 0.725 & 0.806 \\
            \hline
            o1-mini & 0.763 & 0.772 & 0.753 \\
            \hline
            o1-mini + gemini-1.5-flash + gpt-4o + gpt-4o-mini & 0.758 & 0.706 & 0.819 \\
            \hline
            ontology + gpt-4o & 0.758 & 0.790 & 0.728 \\
            \hline
            o1-mini + gemini-1.5-flash + gpt-4o-mini & 0.758 & 0.715 & 0.806 \\
            \hline
            gemini-1.5-flash + ontology + gpt-4o + gpt-4o-mini & 0.754 & 0.742 & 0.766 \\
            \hline
            o1-mini + gemini-1.5-flash & 0.754 & 0.742 & 0.766 \\
            \hline
            o1-mini + ontology + gpt-4o & 0.747 & 0.689 & 0.816 \\
            \hline
            gemini-1.5-flash + gpt-4o-mini & 0.746 & 0.828 & 0.678 \\
            \hline
            ontology + gpt-4o + gpt-4o-mini & 0.744 & 0.744 & 0.744 \\
            \hline
            o1-mini + gemini-1.5-flash + ontology + gpt-4o & 0.744 & 0.683 & 0.816 \\
            \hline
            o1-mini + ontology + gpt-4o + gpt-4o-mini & 0.738 & 0.668 & 0.825 \\
            \hline
            o1-mini + ontology + gpt-4o-mini & 0.738 & 0.675 & 0.813 \\
            \hline
            o1-mini + gemini-1.5-flash + ontology + gpt-4o + gpt-4o-mini & 0.735 & 0.663 & 0.825 \\
            \hline
            o1-mini + gemini-1.5-flash + ontology + gpt-4o-mini & 0.734 & 0.670 & 0.813 \\
            \hline
            o1-mini + ontology & 0.731 & 0.694 & 0.772 \\
            \hline
            gemini-1.5-flash + ontology + gpt-4o-mini & 0.730 & 0.756 & 0.706 \\
            \hline
            o1-mini + gemini-1.5-flash + ontology & 0.728 & 0.688 & 0.772 \\
            \hline
            gemini-1.5-flash & 0.710 & 0.903 & 0.584 \\
            \hline
            gemini-1.5-flash + ontology & 0.697 & 0.798 & 0.619 \\
            \hline
            ontology + gpt-4o-mini & 0.648 & 0.728 & 0.584 \\
            \hline
            gpt-4o-mini & 0.598 & 0.797 & 0.478 \\
            \hline
            ontology & 0.480 & 0.723 & 0.359 \\
            \hline
    \end{tabular}
  \end{center}
  \caption{Performance metrics (F1-score, precision, and recall) for all model configurations evaluated on the Symptom Extraction task, sorted by F1-score from highest to lowest.}\label{tab:SymptomComparison}
\end{table}

\begin{table}[H]
  \begin{center}
    \begin{tabular}[c]{|l|l|l|l|l|l|l|}
      \hline
      \textbf{Task} & \textbf{Model} & \textbf{F1-Score} & \textbf{NPV} & \textbf{PPV} & \textbf{Sensitivity} & \textbf{Specificity} \\
      \hline
      Disease Extraction & BaseBERT & 0.830 & 0.990 & 0.785 & 0.890 & 0.980 \\
      & BioBERT & 0.865 & 0.990 & 0.825 & 0.910 & 0.985 \\
      & BioClinBERT & 0.830 & 0.990 & 0.780 & 0.890 & 0.975 \\
      \hline
      Medication Extraction & BaseBERT & 0.885 & 1.000 & 0.885 & 0.885 & 1.000 \\
      & BioBERT & 0.890 & 1.000 & 0.890 & 0.890 & 1.000 \\
      & BioClinBERT & 0.890 & 1.000 & 0.895 & 0.890 & 1.000 \\
      \hline
      Symptom Extraction & BaseBERT & 0.330 & 0.985 & 0.365 & 0.295 & 1.000 \\
      & BioBERT & 0.340 & 0.985 & 0.400 & 0.295 & 1.000 \\
      & BioClinBERT & 0.340 & 0.985 & 0.385 & 0.305 & 1.000 \\
      \hline
    \end{tabular}
  \end{center}
  \caption{Performance of BaseBERT, BioBERT, and BioClinBERT for each NER task, averaged across teacher labels and human labels.}\label{tab:BERTComparison}
\end{table}

\begin{table}[H]
  \begin{center}
    \begin{tabular}[c]{|l|l|l|l|}
      \hline
      \textbf{Task} & \textbf{Precision (\%)} & \textbf{Recall (\%)} & \textbf{F1 (\%)} \\
      \hline
      Disease & 61.3 & 89.0 & 72.6 \\
      Medication & 81.5 & 96.4 & 88.3 \\
      Symptom & 92.9 & 56.0 & 69.9 \\
      \hline
    \end{tabular}
  \end{center}
  \caption{Performance of the best performing distilled BioBERT model on the external validation dataset.}\label{tab:ExternalValidation}
\end{table}

\begin{table}[H]
  \begin{center}
    \begin{tabular}[c]{|l|l|l|l|l|l|l|l|l|}
      \hline
      \multicolumn{4}{|c|}{} & \multicolumn{3}{c}{Tokens per Document} & \multicolumn{2}{|c|}{Labels} \\
      \hline
      \textbf{Dataset} & \textbf{Task} & \textbf{Split} & \textbf{N} & \textbf{Min} & \textbf{Median} & \textbf{Max} & \textbf{O} & \textbf{Entity} \\
      \hline
      NCBI & Disease & Train & 593 & 36 & 202 & 555 & 98637 & 7507 \\
      & & Test & 100 & 72 & 203.5 & 487 & 17832 & 1385 \\
      & & Dev & 100 & 83 & 202.5 & 366 & 17724 & 1185 \\
      \hline
      n2c2 & Medication & Train & 303 & 159 & 2479 & 9285 & 642067 & 18099 \\
      & & Test & 202 & 191 & 2576 & 8705 & 419779 & 11995 \\
      & & Dev & 25 & 168 & 2560 & 5404 & 54512 & 1553 \\
      \hline
      CORAL & Symptom & Train & 200 & 631 & 2224 & 6818 & - & - \\
      & & Test & 35 & 1087 & 2374 & 4939 & 78887 & 1402 \\
      & & Dev & 5 & 1762 & 2509 & 3167 & 10586 & 274 \\
      \hline
      MIMIC-III & - & Train & 1000 & 42 & 315.5 & 3008 & - & - \\
      \hline
      Medalign & All & Test & 746 & 20 & 378.5 & 3047 & - & - \\
      \hline
    \end{tabular}
  \end{center}
    \caption{Summary statistics for all clinical document datasets, including document counts (N), token counts per document, and entity label counts based on ground truth human labels. Development splits were used for prompt engineering, training splits were used for generating teacher labels, and testing splits were used for evaluation. Abbreviations: “O” = “Outside” (indicating non-entity).}\label{tab:DataOverview}
\end{table}

\begin{table}[H]
  \begin{center}
    \begin{tabular}[c]{|l|l|}
      \hline
      \textbf{Task} & \textbf{Cohen's Kappa} \\
      \hline
      NCBI & 0.88 \\
      n2c2 & 0.86 \\
      CORAL & 0.67 \\
      MedAlign & 0.61 \\
      \hline
    \end{tabular}
  \end{center}
  \caption{Inter-rater reliability results for error analysis and external validation (MedAlign).  Cohen’s Kappa for MedAlign is a multiclass metric.}\label{tab:InterRater}
\end{table}

\subsection*{Prompts}

The following prompts yielded the highest F1 score on the development sets for each task. GPT-4o was used for all prompt tuning.

\subsubsection*{Medication Extraction}

\noindent\fbox{%
    \parbox{\textwidth}{%
List all medications, drugs, and drug classes mentioned in the following clinical note. Illicit drugs and alcohol should not be listed.
\begin{enumerate}
    \item Make sure to include:
    \begin{enumerate}
        \item Specific medication names (both brand and generic).
        \begin{itemize}
            \item If present in the note, include the full name, like tiotropium bromide or albuterol sulfate, instead of just tiotropium or albuterol.
        \end{itemize}
        \item Drug class names, both singular and plural, including (but not limited to):
        \begin{itemize}
            \item NSAID, anticoagulant, pain medication, PPI, steroids, antibiotics, ACE inhibitors, pressors, sedating medications, etc.
        \end{itemize}
        \item Substances that are injected or infused, such as:
        \begin{itemize}
            \item Fluids, iron, contrast dye, red blood cells (pRBC), platelets, etc.
        \end{itemize}
        \item Substances that are inhaled, such as:
        \begin{itemize}
            \item Oxygen, FIO2, nebulized medications, inhaled bronchodilators, etc.
        \end{itemize}
    \end{enumerate}

    \item Do not include:
    \begin{enumerate}
        \item Medical devices or equipment, such as:
        \begin{itemize}
            \item Inhaler, nebulizer, BiPAP machine, etc.
        \end{itemize}
        \item Modes of administration or formulations, such as:
        \begin{itemize}
            \item IV, drip, gtt (drops), liquid form, transfused, supplement, etc.
        \end{itemize}
        \item Methods of delivery or routes of administration, unless part of the medication's name.
        \item General descriptors or measurements (e.g., units, mg, ml, 80, 30) unless these are part of a medication's name.
    \end{enumerate}

    \item Additional Notes:
    \begin{itemize}
        \item Avoid listing terms like increased, solution (Soln), isotonic, or daycare, which are not medications or drug classes.
        \item Include only pharmacologically relevant terms (e.g., antibiotic, anticoagulant, steroid).
    \end{itemize}

    \item Examples of Challenging Cases:
    \begin{enumerate}
        \item False Positives to Avoid:
        \begin{itemize}
            \item Oxygen as a standalone word unless explicitly used as a therapy or treatment.
            \item Do not list FIO2 unless explicitly described as oxygen therapy.
            \item Avoid terms like isotonic or Sodium unless part of a drug name (e.g., heparin sodium).
        \end{itemize}
        \item False Negatives to Include:
        \begin{itemize}
            \item Drug classes (e.g., steroids, antibiotics, ACE inhibitors).
            \item Medications with brand or generic names (e.g., albuterol, tiotropium bromide, Zithromax).
            \item Injectable medications (e.g., ceftriaxone, heparin).
                \newline
                \newline
        \end{itemize}
    \end{enumerate}
\end{enumerate}
Output Format
Output a string delimited by // to separate the medications. Write them exactly as they were written in the note. Do not output anything else.
\newline
\newline
Example:
Note:  
"Patient is experiencing muscle pain, secondary to statin therapy for coronary artery disease.
The patient suffers from steroid-induced hyperglycemia. Patient prescribed 1 x 20 mg Prednisone tablet daily for 5 days.
Patient has been switched to lisinopril tablet 10mg 1 tablet PO QD. Patient received 100 Units/kg IV
heparin sodium injection for treatment of deep vein thrombosis. Sulfa (sulfonamide antibiotics).
Tylenol (Acetaminophen) B.i.d. (twice a day)."
\newline
\newline
Output:  
\{\{"entities": "statin // steroid // Prednisone // lisinopril // heparin sodium // Sulfa (sulfonamide antibiotics) // Tylenol (Acetaminophen)",
"rationale": "All medication names, including drug classes and drug names in parentheses, were extracted. Dosages (eg. 20mg) and other administration information (eg. injection) were not extracted as per the instructions."\}\}
\newline
\newline
Here is the note:
\{note\}
    }%
}

\subsubsection*{Symptom Extraction}

\noindent\fbox{%
    \parbox{\textwidth}{%
List all symptoms explicitly mentioned in the following clinical note. 
\begin{enumerate}
    \item Include:
    \begin{itemize}
        \item All mentions of symptoms or complaints, such as "fatigue," "nausea," "vomiting," or "pain."
        \item Negated symptoms (e.g., "denies nausea" should still include "nausea").
        \item Minimal entity spans: Use the simplest terms that convey the symptom information (e.g., "nausea" instead of "the patient presents with nausea").
        \item Severity modifiers, when relevant (e.g., "minor fatigue," "severe pain").
    \end{itemize}

    \item Do not include:
    \begin{itemize}
        \item Signs or clinical findings that are not symptoms (e.g., "hyperbilirubinemia," "ascites").
        \item Information about the location of the symptom (e.g., "low back pain" $\rightarrow$ Include only "pain").
        \item Adjectives or descriptors unrelated to the symptom itself (e.g., "low," "new," "right-sided").
        \item Conjunctions, prepositions, or other grammatical words unrelated to the symptoms (e.g., "and," "of," "at").
    \end{itemize}

    \item Edge Cases:
    \begin{itemize}
        \item For combinations of symptoms (e.g., "nausea and vomiting"), list each symptom separately (e.g., "nausea // vomiting").
        \item Avoid listing any context or causes of the symptom. For example:
        \begin{itemize}
            \item "pain secondary to surgery" $\rightarrow$ Include only "pain."
            \item "headache from dehydration" $\rightarrow$ Include only "headache."
        \end{itemize}
        \item Make sure to list any symptoms mentioned, even those that the patient is negative for.
        \newline
        \newline
    \end{itemize}

\end{enumerate}
Output Format
Output a JSON with two keys:
\begin{enumerate}
    \item "entities": a single string of symptoms, separated by `//`. Write the symptoms exactly as they appear in the note.
    \item "negated\_symptoms\_included": a string, affirming that all negated symptoms were included
    \newline
    \newline
\end{enumerate}
Example:
Note:
"Patient is experiencing muscle pain, lower back pain, and fatigue,  
secondary to statin therapy for coronary artery disease. Patient denies nausea and vomiting.  
The patient suffers from steroid-induced hyperglycemia. Negative for fever, weight loss, and dysuria.  
Patient prescribed 1 x 20 mg Prednisone tablet daily for 5 days.  
Patient has been switched to lisinopril tablet 10mg 1 tablet PO QD.  
Patient received 100 Units/kg IV heparin sodium injection for treatment of deep vein thrombosis.  
Sulfa (sulfonamide antibiotics). Tylenol (Acetaminophen) B.i.d. (twice a day)."
\newline
\newline
Output:
{{"entities": "pain // fatigue // nausea // vomiting // fever // weight loss // dysuria",  
"negated_symptoms_included": "Yes, all negated symptoms were listed (nausea, vomiting, fever, weight loss, dysuria)"}}
\newline
\newline
Here is the note:
\{note\}
    }%
}

\subsubsection*{Disease Extraction}

\noindent\fbox{%
    \parbox{\textwidth}{%
List all diseases, disorders, and clinical conditions mentioned in the following clinical note.
\begin{enumerate}
    \item Include:
    \begin{itemize}
        \item Specific diseases and disorders
        \begin{itemize}
            \item Examples: "ataxia-telangiectasia", "Phenylketonuria", "Aniridia"
        \end{itemize}
        \item Disease categories or classes
        \begin{itemize}
            \item Examples: "Inherited human disease", "Chromosome abnormalities", "Cancer"
        \end{itemize}
        \item Composite mentions indicating diseases or conditions
        \begin{itemize}
            \item Examples: "Combined deficiency of C6 and C7", "Stage II colorectal carcinoma", "Segmental necrotizing glomerulonephritis"
        \end{itemize}
        \item Modifiers that describe diseases or conditions
        \begin{itemize}
            \item Examples: "Deficiency of hepatic phenylalanine hydroxylase", "Myocardial lesions", "DCC-negative tumors"
        \end{itemize}
        \item Abbreviations or acronyms referring to diseases or conditions
        \begin{itemize}
            \item Examples: "A-T" for ataxia-telangiectasia, "HD" for Huntington's disease
        \end{itemize}
        \item Plural forms and variations of disease terms
        \begin{itemize}
            \item Examples: "Lipomas", "Cancers", "Tumors", "Pleural effusions"
        \end{itemize}
        \item Symptoms, signs, and clinical findings
        \begin{itemize}
            \item Examples: "Bradycardia", "Hypotension", "Pleural effusion", "Tonic-clonic seizures", "Dyspnoea", "Chest pain", "Fever", "Neurologic impairment", "Fatiguability", "Dizziness", "Syncopal attacks", "Blindness", "Eosinophilia", "Urinary abnormalities", "Red eyes", "Asystolic", "Overdose", "Toxicity", "Stable angina"
        \end{itemize}
    \end{itemize}

    \item Exclude:
    \begin{itemize}
        \item Genetic mutations or hypotheses without explicit disease mention
        \begin{itemize}
            \item Examples: "Disease-causing mutations", "Hypothesis of a defective gene"
        \end{itemize}
        \item Hypothetical or unconfirmed conditions
        \begin{itemize}
            \item Examples: "Hypothesis of a defective C9", "Compound heterozygote for uncharacterized genes"
        \end{itemize}
        \item Traits or responses not specifying a disease
        \begin{itemize}
            \item Examples: "Radio-sensitive phenotype", "Defective cell cycle checkpoints"
        \end{itemize}
        \item Descriptions of biological processes or impairments not representing a specific disease
        \begin{itemize}
            \item Examples: "Functional impairment", "T-cell-dependent immune responses", "Secretion abnormalities"
        \end{itemize}
        \item General observations or modifiers
        \begin{itemize}
            \item Examples: "Reduced immune function", "Impaired secretion"
        \end{itemize}
        \item Broad functional or descriptive terms unless tied directly to a disease
        \begin{itemize}
            \item Examples: "Impairment", "Deficiency" (unless part of a recognized condition like "T-cell deficiency")
        \end{itemize}
    \end{itemize}

    \item Additional Instructions for Acronyms:
    \begin{itemize}
        \item Focus on identifying acronyms that represent diseases, disorders, and findings. Ensure no acronyms are omitted from the output.
        \newline
        \newline
    \end{itemize}
\end{enumerate}

Output Format
Output a string delimited by `//` to separate the diseases. Write them exactly as they were written in the note. Do not output anything else.
\newline
\newline
Here is the note:
\{note\}

   }%
}

\end{document}